\documentclass[10pt,twocolumn]{article}

\usepackage{iccv}
\usepackage{times}
\usepackage{epsfig}
\usepackage{graphicx}
\usepackage{amsmath}
\usepackage{amsthm}
\usepackage{amssymb}
\usepackage{multirow}
\usepackage{footmisc}
\usepackage{color}
\usepackage{threeparttable}
\usepackage{soul}
\usepackage{nopageno}

\usepackage[pagebackref=true,breaklinks=true,colorlinks,bookmarks=false]{hyperref}

\iccvfinalcopy % *** Uncomment this line for the final submission

\def\iccvPaperID{4330} % *** Enter the ICCV Paper ID here

\ificcvfinal\fi
\begin{document}
\thispagestyle{empty}
%%%%%%%%% TITLE
\title{ABD-Net: Attentive but Diverse Person Re-Identification}

\author{Tianlong Chen\textsuperscript{1}, Shaojin Ding\textsuperscript{1}\footnotemark[1] , Jingyi Xie\textsuperscript{2}\footnotemark[1] , Ye Yuan\textsuperscript{1}, Wuyang Chen\textsuperscript{1} \\ Yang Yang\textsuperscript{3}, Zhou Ren\textsuperscript{4}, Zhangyang Wang\textsuperscript{1}\\
\textsuperscript{1}Texas A\&M University, \textsuperscript{2}University of Science and Technology of China \\ \textsuperscript{3}Walmart Technology, \textsuperscript{4}Wormpex AI Research\\
\textit{\small \{wiwjp619,shjd,ye.yuan,wuyang.chen,atlaswang\}@tamu.edu}\\
\textit{\small hsfzxjy@mail.ustc.edu.cn}, 
\textit{\small yang.yang2@walmart.com}, \textit{\small zhou.ren@bianlifeng.com}\\
\url{https://github.com/TAMU-VITA/ABD-Net}
}
% \{wiwjp619,dshj940428,ye.yuan,wuyang.chen,atlaswang\}@tamu.edu \\ hsfzxjy@mail.ustc.edu.cn, yang.yang2@walmart.com, renzhou200622@gmail.com\\}

% \and

\maketitle
%\thispagestyle{empty}

%%%%%%%%% ABSTRACT
\begin{abstract}
   Attention mechanisms have been found effective for person re-identification (Re-ID). However, the learned ``attentive'' features are often not naturally uncorrelated or ``diverse'', which compromises the retrieval performance based on the Euclidean distance. We advocate the complementary powers of attention and diversity for Re-ID, by proposing an Attentive but Diverse Network (ABD-Net). ABD-Net seamlessly integrates attention modules and diversity regularizations throughout the entire network to learn features that are representative, robust, and more discriminative. Specifically, we introduce a pair of complementary attention modules, focusing on channel aggregation and position awareness, respectively. Then, we plug in a novel orthogonality constraint that efficiently enforces diversity on both hidden activations and weights. Through an extensive set of ablation study, we verify that the attentive and diverse terms each contributes to the performance boosts of ABD-Net. It consistently outperforms existing state-of-the-art methods on there popular person Re-ID benchmarks.
   
    %However, the outputs of conventional attention mechanisms are often highly correlated and redundant. As a result, the power of attention mechanism will remain limited, and thus jeopardize person Re-ID performance. In this paper, we propose an Attentive but Diverse Network (ABD-Net) to promote not only informativeness but also the diversity of the learned features, and therefore the learned features can span a more representative feature space. ABD-Net consists of an attention module and an orthogonality regularization. The attention module extracts local patterns for person region to form an informative feature embedding \textcolor{red}{[wuyang] I'm not sure if using the word ``local'' here is proper since attention is a non-local operation. I re-phrased this line ``Non-local but similar patterns for person region are aggregated by the attention module to form an informative feature embedding''}. Further regularization of orthogonality on both features and weights enforces the diversity of the feature embedding. Extensive experiments are conducted on three benchmark datasets, consistently and significantly demonstrated the superior performance of our ABD-Net comparing to existing state-of-the-art methods.
\end{abstract}
\vspace{-5mm}

\renewcommand{\thefootnote}{\fnsymbol{footnote}}
\footnotetext[1]{Equal Contribution.}
%\footnotetext[2]{Zhangyang Wang is the corresponding author.}

%%%%%%%%% BODY TEXT
\section{Introduction}

Person Re-Identification (Re-ID) aims to associate individual identities across different time and locations. It embraces many applications in intelligent video surveillance. Given a query image and a large set of gallery images, person Re-ID represents each image with a \textit{feature embedding}, and then ranks the gallery images in terms of feature embeddings' similarities to the query. Despite the exciting progress in recent years, person Re-ID remains to be extremely challenging in practical unconstrained scenarios. Common challenges arise from body misalignment, occlusion, background perturbance, view point changes, pose variations and noisy labels, among many others \cite{Wei_2018_CVPR}.

Substantial efforts have been devoted to addressing those various challenges. Among them, incorporating body part information \cite{gray2008viewpoint, prosser2010person, cheng2016person, su2017pose, zheng2017pose} has empirically proven to be effective in enhancing the feature robustness against body misalignment, incomplete parts, and occlusions. Motivated by such observations, the attention mechanism \cite{vaswani2017attention} was introduced to enforce the features to mainly capture the discriminative appearances of human bodies (or certain body parts). Since then, the attention-based models \cite{zhao2017deeply, yao2017deep, si2018dual, xu2018attention, li2018harmonious} have boosted person Re-ID performance much. 

\begin{figure}[t]
\begin{center}
   \includegraphics[width=1\linewidth]{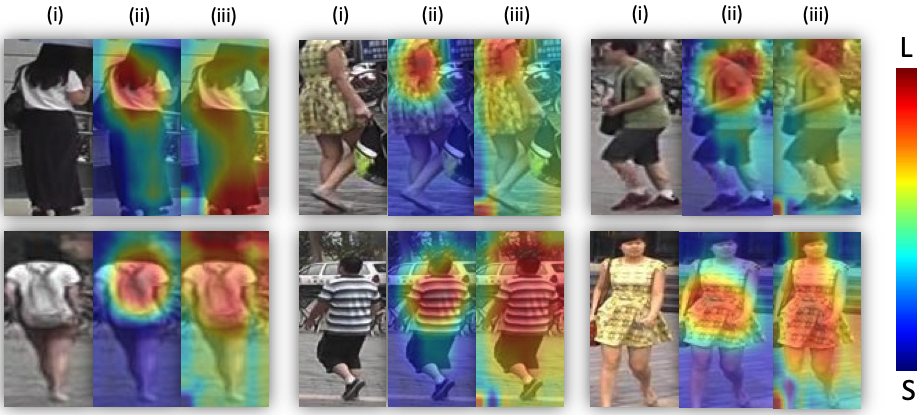}
\end{center}
   \caption{Visualization of attention maps. (i) Original images; (ii) Attentive feature maps; (iii) Attentive but diverse feature maps. In general, diversity is observed to make attention ``broader'' and to reduce the (incorrect) overfitting of local regions (such as clothes textures) by attention. (L: large values; S: small values)}
\label{fig:visfintro}
\vspace{-5mm}
\end{figure}

On a separate note, the feature embeddings are used to compute similarities between images, typically based on the Euclidean distance, to return the closest matches. Sun et al. \cite{Sun_2017} pointed out that correlations among feature embeddings would significantly compromise the matching performance. The low feature correlation property is, however, not naturally guaranteed by attention-based models. Our observation is that those attention-based models are often more prone to higher feature correlations, because intuitively, the attention mechanism tends to have features focus on a more compact subspace (such as foreground instead of the full image, see Fig.～\ref{fig:visfintro} for examples). %(\textcolor{red}{More detailed visualization results can be found in Section 4.4 TBD}). 

In view of the above, we argue that a more desirable feature embedding for person Re-ID should be both \textbf{attentive} and \textbf{diverse}: the former aims to correct misalignment, eliminate background perturbance, and focus on discriminative local parts of body appearances; the latter aims to encourage lower correlation between features, and therefore better matching, and potentially make feature space more comprehensive. We propose an Attentive but Diverse Network (\textbf{ABD-Net}), that strives to integrate attention modules and diversity regularization and enforces them throughout the entire network. The main contributions of ABD-Net are outlined as below:
\begin{itemize}
\vspace{-0.3em}
    \item We incorporate a compound attention mechanism into ABD-Net, consisting of \textit{Channel Attention Module} (CAM) and \textit{Position Attention Module} (PAM). CAM facilitates channel-wise, feature-level information aggregation, while PAM captures the spatial awareness of body and part positions. They are found to be complementary and altogether benefit Re-ID.\vspace{-0.5em}
    \item We introduce a novel regularization term called \textit{spectral value difference orthogonality} (SVDO) that directly constrains the conditional number of the weight Gram matrix. SVDO, efficiently implemented, is applied to both activations and weights, and is shown to effectively reduce learned feature correlations.\vspace{-0.5em}
    \item We perform extensive experiments on Market-1501 \cite{Zheng_2015_ICCV}, DukeMTMC-Re-ID \cite{Ristani_2016_ECCV}, and MSMT17 \cite{Wei_2018_CVPR}. ABD-Net significantly outperforms existing methods,  achieving new state-of-the-art on all three popular benchmarks. We also verify that the attentive and diverse terms each contributes to a performance gain, through rigorous ablation studies and visualizations.\vspace{-0.3em}
\end{itemize}
\section{Related Work} \label{sec:rw}

\subsection{Person Re-identification: Brief Overview} %\textcolor{red}{[wuyang] I'm not sure if using a more specific section name like ``Effective feature extraction in person re-identification'' would be better, since solely ``Person re-identification'' makes me feel too board and not specific enough according to the problem you want to solve. Anyway I'm not an expert in re-id and please talk to Prof. Wang.} \textcolor{blue}{[Tianlong] I think it is a good point. But this part is not about effective feature extraction. It talks about the original person-Re-ID problem.}
Person Re-ID has two key steps: obtaining a feature embedding and performing matching under some distance metric \cite{weinberger2006distance, li2013learning, khamis2014joint}. We mainly review the former where both handcrafted features \cite{khamis2014joint, koestinger2012large, li2013locally, ma2012bicov} and learned features \cite{li2014deepreid, zhao2014learning, cheng2016person, hermans2017defense, liao2017triplet} were studied. 
%For the latter, various efforts \cite{weinberger2006distance, li2013learning, khamis2014joint} were also adopted. 
In recent years, the prevailing success of convolutional neural networks (CNNs) in computer vision has made person Re-ID no exception. Due to many problem-specific challenges such as occlusion/misalignment, incomplete body parts, as well as background perturbance/view point changes, 
naively applying CNN backbones to feature extraction may not yield ideal Re-ID performance. Both image-level features and local features extracted from body parts prove to enhance the robustness. Many part-based methods have achieved superior performance \cite{gray2008viewpoint, prosser2010person, liao2015person, ma2013domain, zheng2013reidentification, cheng2016person, su2017pose, wei2017glad, zheng2017pose, zhu2017part, suh2018part, zhao2017deeply, zhao2017spindle}. We refer readers to \cite{zheng2016person} for a more comprehensive review.

%To obtain more effective representations for person images, both conventional handcrafted features \cite{khamis2014joint, koestinger2012large, li2013locally, ma2012bicov} and deep learning features \cite{li2014deepreid, zhao2014learning, cheng2016person, hermans2017defense, liao2017triplet} were explored. With feature representations extracted, distance metrics \cite{weinberger2006distance, li2013learning, khamis2014joint} can hence be applied to make a final decision.

%\textbf{Deep learned features for person Re-ID.} Noticing the great success of deep convolutional neural network (DCNN) in various computer vision tasks, an increasing number of studies used DCNN to learn the feature embedding in person Re-ID. One of the most challenging problems in DCNN based person Re-ID is the alignments between people from different images. 

%Due to the occlusion, incomplete body parts, visually similar backgrounds, and view point changes, the alignments are often error-prone. To enhance the robustness of the alignments, both global features extracted from the entire image and local features extracted from body parts are jointly considered. In these part-based methods, both hand-crafted part features \cite{gray2008viewpoint, prosser2010person, liao2015person, ma2013domain, zheng2013reidentification} and DCNN learned part features \cite{cheng2016person, su2017pose, wei2017glad, zheng2017pose, zhu2017part, suh2018part, zhao2017deeply, zhao2017spindle} were explored, with concentration shifted to the latter.

\subsection{Attention Mechanisms in Person Re-ID}
Several studies proposed to integrate attention mechanism into deep models to address the misalignment issue in person Re-ID. Zhao et al. \cite{zhao2017deeply} proposed a part-aligned representation based on a part map detector for each predefined body part. Yao et al. \cite{yao2017deep} proposed a Part Loss Network which defined a loss for each average pooled body part and jointly optimized the summation losses. Si et al. \cite{si2018dual} proposed a dual attention matching network based on an inter-class and an intra-class attention module to capture the context information of video sequences for person Re-ID. Li et al. \cite{li2018harmonious} proposed a multi-task learning model that learns hard region-level and soft pixel-level attention jointly to produce more discriminative feature representations. Xu et al. \cite{xu2018attention} used pose information to learn attention masks for rigid and non-rigid parts, and then combined the global and part features as the final feature embedding.

Our proposed attention mechanism differs from previous methods in several aspects. First, previous methods \cite{zhao2017deeply, yao2017deep, xu2018attention} only use attention mechanisms to extract part-based
%\textcolor{red}{[wuyang] again, the word ``local'', same worry as I mentioned in the abstract} \textcolor{blue}{[Tianlong] Yes. It does appear in several places.} 
spatial patterns from person images, which are usually focus in the foregrounds. In contrast, ABD-Net combines spatial and channel clues; besides, our added diversity constraint will avoid the overly correlated and redundant attentive features.
%also uses the orthogonality regularization to avoid the attentive features to be correlated and redundant.
Second, our attention masks are directly learned from the data and context, without relying on manually-defined parts, part region proposals, nor pose estimation \cite{zhao2017deeply, yao2017deep, xu2018attention}. 
%while \cite{zhao2017deeply, yao2017deep, xu2018attention} hinge on the provided or pre-defined masks of body parts. 
Our two attention modules are embedded within a single backbone, making our model lighter-weight than the multi-task learning alternatives \cite{xu2018attention,li2018harmonious}.

%Third, our attentive but diverse mechanism focuses on people alignment from inaccurate bounding boxes, which is different from \cite{si2018dual} whose attention matching network is to increase the discrimination between different people. Finally, our proposed attentive but diverse module is within a single branch. Comparing with the multi-task learning model proposed in \cite{li2018harmonious}, the single branch model has less parameter, and therefore it is more computationally efficient.

\subsection{Diversity via Orthogonality}
Orthogonality has been widely explored in deep learning to encourage the learning of informative and diverse features. In CNNs, %orthogonality imposed on weights is recognized to stabilize the layer-wise distribution of activations \cite{rodrguez2016regularizing} and make optimization more efficient. 
several studies \cite{DBLP:journals/corr/abs-1102-1523,harandi2016generalized,ozay2016optimization,huang2017orthogonal} perform regularization using ``hard orthogonality constraints", which typically depends on singular value decomposition (SVD) to strictly constrain their solutions on a Stiefel manifold. The similar idea was first exploited by \cite{Sun_2017} for person Re-ID, where the authors performed SVD on the weight matrix of the last layer, in an effort to reduce feature correlations. Despite the effectiveness, SVD-based hard orthogonality constraints are computationally expensive, and sometimes appear to limit the learning flexibility.  
%and is often too expensive for training. In \cite{Sun_2017}, SVDNet performs SVD on the weight matrix from FC layer. It is of vital importance to reduce the redundancy in the FC descriptor to make it work under the Euclidean distance. 
%However, SVDNet is resource consuming in training but does not directly benefit the feature embedding extraction process. 

Recent studies also investigated ``softer" orthogonality regularizations by enforcing the Gram matrix of each weight matrix to be close to an identity matrix, under Frobenius norm \cite{Xie_2017} or spectral norm \cite{bansal2018can}. %Both methods can lead to smaller correlations among learned features but are still too strong to apply.
We propose a novel spectral value difference orthogonality (SVDO) regularization that directly constrains the conditional number of the Gram matrix. Also contrasting from \cite{Sun_2017,Xie_2017} that apply orthogonality only to CNN weights, we enforce the new regularization on both hidden activations and weights.

\section{Attentive but Diverse Network} \label{sec:method}

In this section, we first introduce the two attention modules, followed by the new diversity (orthogonality) regularization. We then wrap them up and describe the overall architecture of ABD-Net.

%\textcolor{red}{The full configurations for CAM and PAM can be found in the supplementary. (I strongly suggest so because details esp. PAM are very clear to me.)} 

%in our Attentive but Diverse network (ABD-Net) with details. Following this, we describe the overall architecture of ABD-Net.

\subsection{Attention: Channel-Wise and Position-Wise}

The goal of attention for Re-ID is to focus on person-related features while eliminating irrelevant backgrounds. Inspired by the successful idea in segmentation \cite{fu2018dual}, we integrate two complementary attention mechanisms: Channel Attention Module (\textbf{CAM}) and Positional Attention Module (\textbf{PAM}). The full configurations for CAM and PAM can be found in the supplementary.

%To learn informative features of people, we design a Channel Attention Module (CAM) and a Positional Attention Module (PAM). The use of CAM and PAM are inspired by their great success in the image segmentation task \cite{fu2018dual}. We further adopt these two modules to be more flexibly used across the whole network for person Re-ID.

\subsubsection{Channel Attention Module} \label{sec:cam}

The high-level convolutional channels in a trained CNN classifier are well-known to be semantic-related and often category-selective. In the person Re-ID case, we hypothesize that the high-level channels in the person Re-ID case are also ``grouped'', \ie, some channels share similar semantic contexts (such as foreground person, occlusions, or background) and are more correlated with each other. CAM is designed to group and aggregate those semantically similar channels. 

%the high-level convolutional layers of a CNN, channels can be regarded as semantic-related features, such as person, occlusions, background in person Re-ID. The basic idea is, if some channel responses share the same semantic context, their correlation will be strong and we could aggregate them in our CAM attention module.

\begin{figure}[t]
\begin{center}
   \includegraphics[width=1\linewidth]{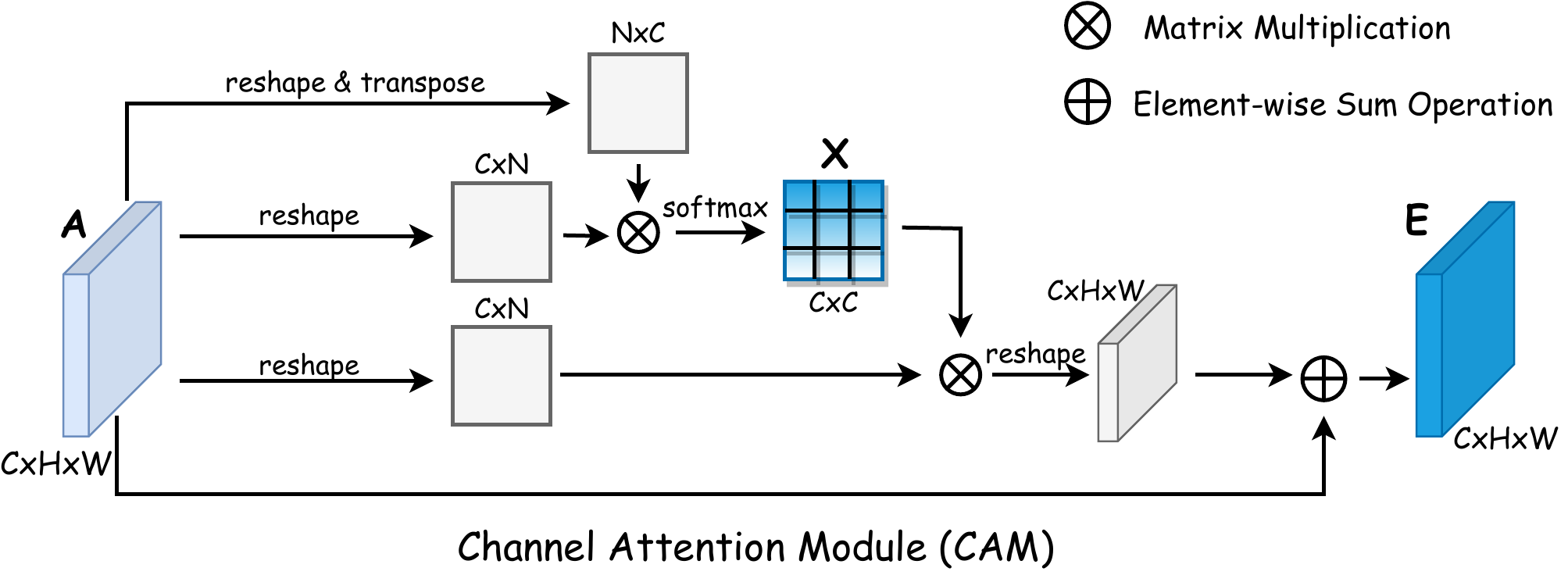}
\end{center}
   \caption{Channel Attention Module (CAM)}
\label{fig:cam}
\vspace{-1mm}
\end{figure}

The full structure of CAM is illustrated in Fig.～\ref{fig:cam}. Given the input feature maps $\textbf{A} \in \mathbb{R}^{C\times H \times W}$, where $C$ is the total number of channels and $H \times W$ is the feature map size, we compute the channel affinity matrix $\textbf{X} \in \mathbb{R}^{C\times C}$, as shown below:
\begin{equation} 
x_{ij}=\frac{exp(A_i\cdot A_j)}{\sum_{j=1}^C exp(A_i\cdot A_j)},\ i,j\in \{1,\cdots,C\}
\label{cam_x}
\end{equation}
where $x_{ij}$ represents the impact of channel $i$ on channel $j$. The final output feature map $\textbf{E}$ is calculated by equation (\ref{cam_e}):

\begin{equation}
E_i = \gamma\sum_{j=1}^C(x_{ij}A_j)+A_i,\ i\in \{1,\cdots,C\}
\label{cam_e}
\end{equation}
$\gamma$ is a hyperparameter to adjust the impact of CAM.

\subsubsection{Position Attention Module} \label{sec:pam}
%Due to local smoothness of images, regions of similar semantic information should have similar pixel values. Namely, pixel values within each semantic context have a larger correlation.

\begin{figure}[t]
\begin{center}
   \includegraphics[width=1\linewidth]{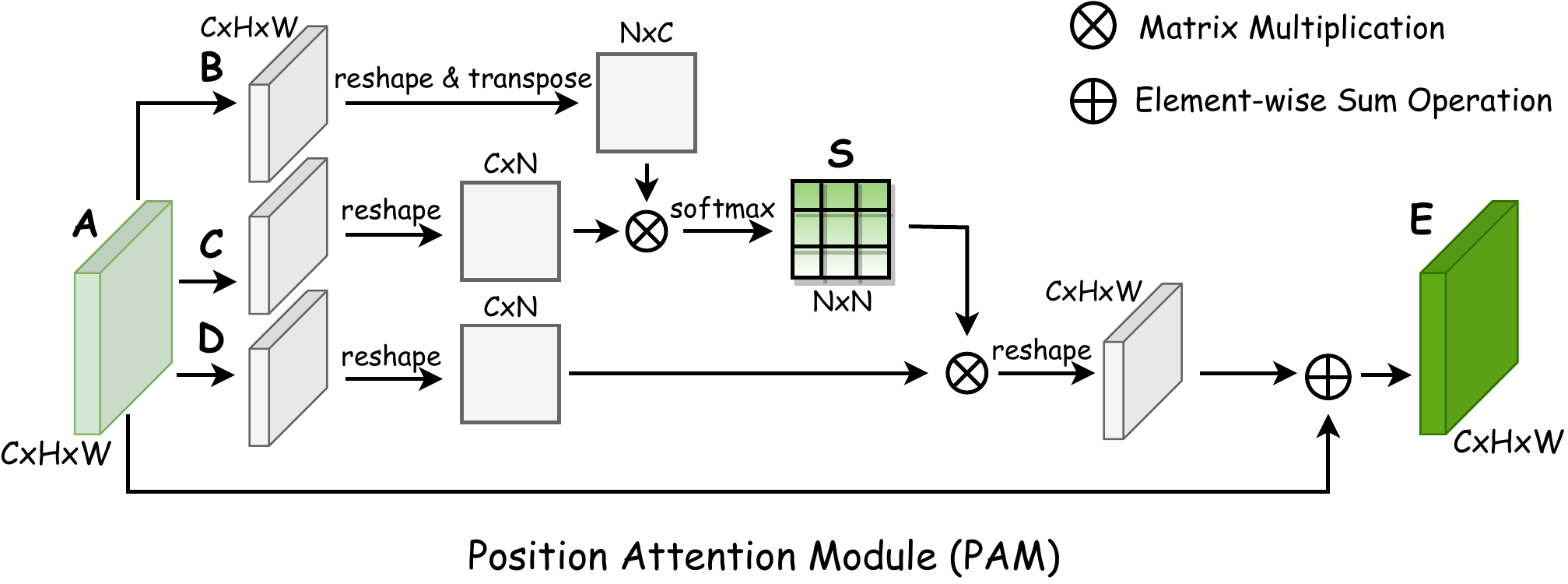}
\end{center}
   \caption{Position Attention Module (PAM)}
\label{fig:pam}
\vspace{-1mm}
\end{figure}

In contrast to CAM, Position Attention Module (PAM) is designed to capture and aggregate those semantically related \textit{pixels} in the spatial domain. We depict the structure of PAM in Fig.～\ref{fig:pam}. The input feature maps $\textbf{A} \in \mathbb{R}^{C\times H \times W}$ are first fed into convolution layers with batch normalization and ReLU activation to produce feature maps $\textbf{B, C, D} \in \mathbb{R}^{C \times H \times W}$. Then we compute the pixel affinity matrix $\textbf{S} \in \mathbb{R}^{N\times N}$ where $N = H \times W$. Note that the dimensions of $S$ and $X$ are different, since the former computes correlations between the total $N$ pixels rather than $C$ channels. We generate the final output feature map $\textbf{E}$ with similar calculation as CAM in Section \ref{sec:cam}.

\begin{figure*}[ht]
\begin{center}
   \includegraphics[width=1\linewidth]{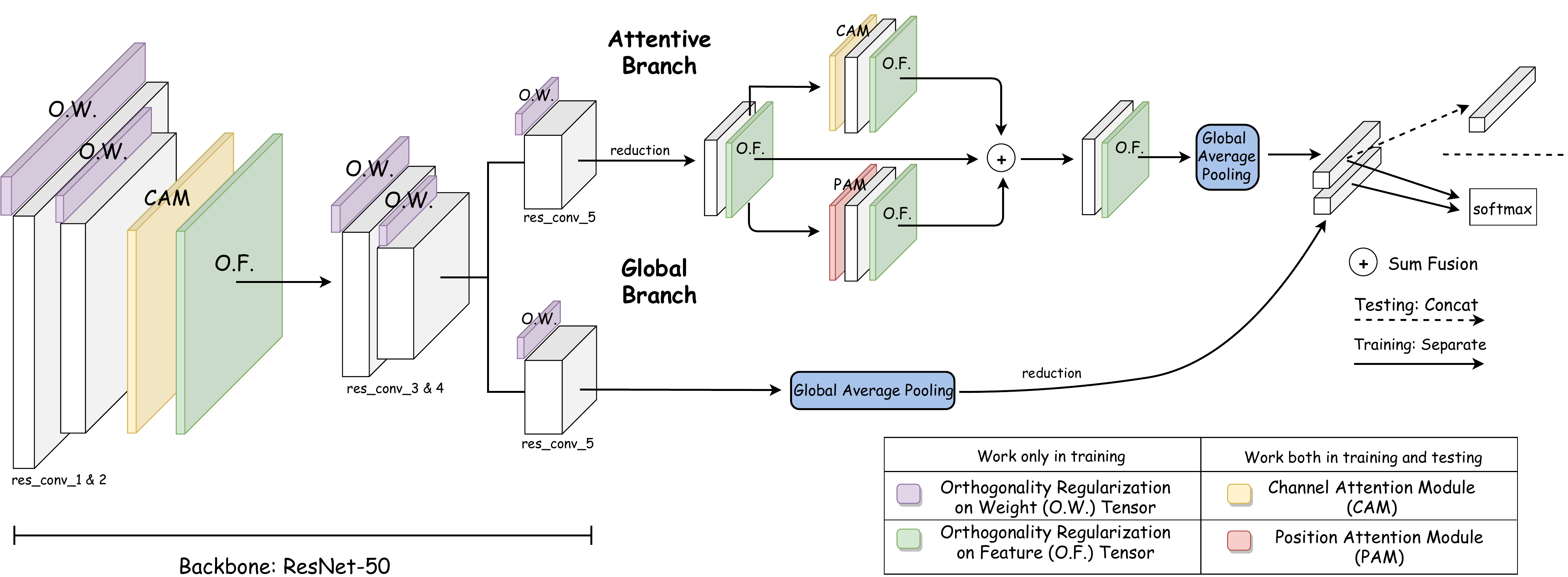}
\end{center}
   \caption{Architecture of ABD-Net: O.W. is applied on all ResNet layers. O.F. is applied after CAM on res\_conv\_2 and after res\_conv\_5 in the Attentive Branch. 
   %res\_conv\_5 in our ABD-Net share a similar structure with original res\_conv\_5, except that they have no down-sampling. 
l   The feature vectors from both attentive and global branches are concatenated as the final feature embedding.}
\label{fig:arch}
\vspace{-1mm}
\end{figure*}

\subsection{Diversity: Orthogonality Regularization}

Following \cite{Sun_2017}, we enforce diversity via orthogonality, yet derive a novel orthogonality regularizer term. It is applied to both hidden features and weights, of both convolutional 
%\textcolor{red}{[wuyang] I did not see you use OF/OW on fully connected layers?} \textcolor{blue}{[Tianlong] You are right, we do not apply the OF/OW on FC, although we can do that. I modify ``is" to ``can be".} 
and fully-connected layers. Orthogonality regularizer on feature space (short for \textbf{O.F.} hereinafter) is to reduce feature correlations that can directly benefit matching. The orthogonal regularizer on weight (\textbf{O.W.}) encourages filter diversity \cite{bansal2018can} and enhances the learning capacity. 

Next, we show the detailed derivation of our orthogonality term on features, while the weight orthogonality can be derived in a similar manner\footnote{For the weight tensor $\textbf{W}_c\in \mathbb{R}^{S\times H\times C\times M}$ in a convolutional layer, where $S,H,C,M$ are filter's width and height, the number of input and output channels, we follow the convention of \cite{Xie_2017,bansal2018can} to reshape $\textbf{W}_c$ into a matrix form $\textbf{F}^*\in \mathbb{R}^{C^*\times M}$, where $C^*=S\times H\times C$. }. For feature maps $\textbf{M}\in \mathbb{R}^{C\times H\times W}$, where $C,H,W$ are the number of channels, feature map's height and width, respectively, we will first reshape $\textbf{M}$ into a matrix form $\textbf{F} \in \mathbb{R}^{C\times N}$, with $N=H\times W$. 
%Our regularization is directly available to almost any CNN. The detailed derivation is described as follows.

%In this section, we derive and describe a new orthogonality regularizer. Note that this regularizer is applicable to both feature space and weight space for both the fully-connected and convolutional layer. 

%The default mathematical expression of the orthogonal regularizer will be assumed on a \textbf{feature or weight} matrix $\textbf{F}\in \mathbb{R}^{m\times n}$.  

Many orthogonality methods  \cite{DBLP:journals/corr/abs-1102-1523,harandi2016generalized,ozay2016optimization,huang2017orthogonal}, including the prior work on person Re-ID \cite{Sun_2017}, enforce hard constraints on orthogonality of weights, whose computations rely on SVD. However, computing SVD on high-dimensional matrices is expensive, urging for the development of soft orthogonality regularizers. Many existing soft regularizers \cite{Xie_2017,xu2019learning} restrict the Gram matrix of $\textbf{F}$ to be close to an identity matrix under Frobenius norm that can avoid the SVD step while being differentiable. However, the gram matrix for an overcomplete $\textbf{F}$ cannot reach identity because of rank deficiency, making those regularizers biased. \cite{bansal2018can} hence introduced the spectral norm-based regularizer that effectively alleviates the bias.

We propose a new option to enforce the orthogonality via  directly regularizing the conditional number of $\textbf{FF}^T$:
\begin{equation}
\beta ||k(\mathbf{F})-1||_2^2\ ,
\label{o_k}
\end{equation}
where $\beta$ is the coefficient and $k(\mathbf{F})$ denotes the condition number of $\textbf{F}$, defined as the ratio of maximum singular value to minimum singular value of $\mathbf{F}$. Naively solving $k(\mathbf{F})$ will take one full SVD. To make it computationally more tractable, we convert (\ref{o_k}) into a spectral value difference orthogonality (SVDO)\footnote{The reason why we choose to penalize the difference between $\lambda_1(\mathbf{FF}^T)$ and $\lambda_2(\mathbf{FF}^T)$ rather than the ratio of them is to avoid numerical instability caused by dividing a very small $\lambda_2(\mathbf{FF}^T)$, which we find happen frequently in our experiments.} regularization:
\begin{equation}
  \beta ||\lambda_1(\mathbf{FF}^T)-\lambda_2(\mathbf{FF}^T)||_2^2\ ,
\label{svdo}
\end{equation}
where $\lambda_1(\mathbf{FF}^T)$ and $\lambda_2(\mathbf{FF}^T)$ denote the largest and smallest eigenvalues of $\textbf{FF}^T$, respectively. 

% We then notice that these two eigenvalues can be derived efficiently using power methods, bypassing the full SVD: 
% \begin{equation}
% \begin{aligned}
%     \lambda_1(FF^T)=&sup_{z\in R^n,z\neq 0}\frac{<A^2z,Az>}{<Az,Az>}\ , \\
%     \lambda_2(FF^T)=&sup_{z\in R^n,z\neq 0}\frac{<B^2z,Bz>}{<Bz,Bz>}+\lambda_1\ .
% \end{aligned}
% \label{lam}
% \end{equation}
% where $\textbf{A=FF}^T$ and $\textbf{B}=(\textbf{FF}^T-\lambda_1\textbf{I})$. 

We use auto-differentiation to obtain the gradient of SVDO, however, this computation still contains the expensive eigenvalue decomposition (EVD). To bypass EVD, we refer to the power iteration method to approximate the eigenvalues. We start with a random initialized $q$, and then iteratively perform equation (\ref{power_m}) (two times by default):
\begin{equation}
%\vspace{-1em}
\begin{aligned}
    p \gets Xq\ ,
    q\gets Xp\ , 
    \lambda(X)\gets \frac{||q||}{||p||}\ .
\end{aligned}
\label{power_m}
\end{equation}
where $\textbf{X}$ in equation (\ref{power_m}) is $\mathbf{FF}^T$ for computing $\lambda_1(\mathbf{FF}^T)$, and $\mathbf{FF}^T-\lambda_1\mathbf{I}$ for $\lambda_2(\mathbf{FF}^T)$. In that way, the computation of SVDO becomes practically efficient.

\subsection{Network Architecture Overview} \label{sec:arc_overview}

%  \textcolor{red}{[wuyang] do you want ot just use the word ``res\_conv\_5'' and avoid using ``res\_conv\_5\_1'' and ``res\_conv\_5\_2'', since I did not see the ``\_1''``\_2'' in the figure and have no idea what's the difference among them. }

%The backbone of ABD-Net can be any network architecture designed for image classification (without fully-connected layers), 

The overall architecture of the proposed ABD-Net is shown in Fig.～\ref{fig:arch}.  ABD-Net is compatible with most common feature extraction backbones, such as ResNet \cite{he2016deep}, InceptionNet \cite{szegedy2017inception}, and Densenet \cite{huang2017densely}. Unless otherwise specified, we use ResNet-50 as the default backbone network due to its popularity in Re-ID \cite{li2018support,zeng2018person,wang2018learning,yang2018local,song2018mask,xu2018attention,Li_2018,Liu_2018}.

We add a CAM and O.F. on the outputs of res\_conv\_2 block. The regularized feature map is used as the input of res\_conv\_3. Next, after the res\_conv\_4 block, the network splits into a \textit{global branch} and an \textit{attentive branch} in parallel. We apply O.W. on all conv layers in our ResNet-50 backbone, $i.e.$, from res\_conv\_1 to res\_conv\_4 and the two res\_conv\_5 in both branches. The outputs of two branches are concatenated as the final feature embedding. 

%the layers is divided into two independent branches, named as \textbf{Global Branch} and \textbf{ABD Branch}. Finally, we concatenate all the outputs from both branches as the final feature vector. 

%Notice that we don't use PAM here. The rationale behind it is that applying PAM to the large feature maps at the early stage is computationally intensive since PAM computes the pixel-wise affinity matrix (as will described in Section \ref{sec:pam}). Thus, for the efficiency in training, we remove PAM in the res\_conv\_2. In what follows, 

The \textit{attentive branch} uses the same res\_conv\_5 layer as that in ResNet-50. The output feature map is then fed into a reduction layer\footnote{A reduction layer consists of a linear layer, batch normalization, ReLU, and dropout. See:  \footnotesize{{\url{https://github.com/KaiyangZhou/deep-person-reid}}}} with O.F. applied yielding a smaller 
%\textcolor{red}{Are you sure those hard-coded numbers are always right for all exp?} \textcolor{blue}{[Tianlong] (In almost all of our experiments, 1024 is better than 512, 2048 maybe better but to expensive. So shall we just hide this detail and put it in sub. materials?) [Shaojin] Just a thought, is it better to put the numbers to implementational details? We only focus on the architecture here rather than all the hyper-parameters} 
feature map $\textbf{T}_a$. We feed $\textbf{T}_a$ into a CAM and a PAM simultaneously, both with O.F. constraints. The outputs from both attentive modules are concatenated with the input $\textbf{T}_a$, and altogether go through a global average pooling layer, ending up with a $k_a$-dimension feature vector. 
%\textcolor{red}{[wuyang] agree with Shaojin: no need to mention specific dim\# here. We can put it in implementation or supple. Same issue in the paragraph below.} \textcolor{blue}{[Tianlong] I read MGN paper, it seems multi-branch model paper like to address the feature dimension.}

In the \textit{global branch}, after res\_conv\_5\footnote{For both two res\_conv\_5 layers in two branches, we removed the down-sampling layer, in order for larger feature maps.}, the feature map $\textbf{T}_g$ is fed into a global average-pooling layer followed by a reduction layer, leading to a $k_g$-dimension feature vector. The global branch intends to preserve global context information in addition to the attentive branch features.

% Practically, we find vertically splitting $T$ into $N_{parts}$ parts result in better performance. We then get $N_{parts}$ 1024-dim feature vectors. In this paper we fix $N_{parts}=2$.

% In the \textbf{Tricky Branch}, we also remove the down-sampling in res\_conv\_5\_2. The $2048 \times 24 \times 8$ output feature map of res\_conv\_5\_2 is firstly reduced into size $1024 \times 24 \times 8$, and then vertically split into two parts. Consequently, we get two feature maps of size $1024 \times 12 \times 8$. For each feature map, we feed it into a dual attention module, and then average pool the output into a 1024-dim feature vector. The two 1024-dim feature vectors are then used for training and testing.

%\subsection{Loss Functions for Training ABD-Net}
Eventually, ABD-Net is trained under the loss function $L$ consisting of a cross entropy loss, a hard mining triplet loss, and orthogonal constraints on feature (O.F.) and on weights (O.W.) penalty terms:
\begin{equation}
L=L_{xent}+\beta_{tr} L_{triplet}+\beta_{O.F.} L_{O.F.}+\beta_{O.W.} L_{O.W.}
\label{loss}
\end{equation}
where $L_{O.F.}$ and $L_{O.W.}$ stand for the SVDO penalty term applied to the hidden features and weights, respectively. $\beta_{tr}, \beta_{O.F.}$ and $\beta_{O.W.}$ are hyper-parameters.

\section{Experiments} \label{sec:exp}
To evaluate ABD-Net, we conducted experiments on three large-scale person re-identification datasets: Market-1501 \cite{Zheng_2015_ICCV}, DukeMTMC-Re-ID \cite{Ristani_2016_ECCV} and MSMT17 \cite{Wei_2018_CVPR}. %\textcolor{blue}{[Shaojin] Do we also need cuhk01/03 since there are more baselines? [Tianlong] Maybe in the sub. materials?} 
First, we report a set of ablation study (mainly on Market-1501 and DukeMTMC-Re-ID) to validate the effectiveness of each component. Second, we compare the performance of ABD-Net against existing state-of-the-art methods on all three datasets. Finally, we provide more visualizations and analysis to illustrate how ABD-Net has achieved its effectiveness. 

\subsection{Datasets}
{\bf Market-1501} \cite{Zheng_2015_ICCV} comprises 32,668 labeled images of 1,501 identities captured by six cameras. Following \cite{Zheng_2015_ICCV}, 12,936 images of 751 identities are used for training, while the rest are used for testing. Among the testing data, the test probe set has 3,368 images of 750 identities. The test gallery set also includes 2,793 additional distractors.

{\bf DukeMTMC-Re-ID} \cite{Ristani_2016_ECCV} contains 36,411 images of 1,812 identities. These images are captured by eight cameras, among which 1,404 identities appear in more than two cameras and 408 identities (distractors) appear in only one camera. The 1,404 identities are randomly divided, with 702 identities for training and the others for testing. In the testing set, one query image for each ID per camera is chosen for the probe set, while all remaining images including distractors are in the gallery. 

{\bf MSMT17} \cite{Wei_2018_CVPR} is the current largest publicly-available person Re-ID dataset. It has 126,441 images of 4,101 identities captured by a 15-camera network (12 outdoor, 3 indoor). We follow the training-testing split of \cite{Wei_2018_CVPR}. The video is collected with different weather conditions at three-time slots (morning, noon, afternoon). All annotations, including camera IDs, weathers and time slots, are available. MSMT17 is \textbf{significantly more challenging} than the other two, due to its massive scale, more complex and dynamic scenes. Additionally, the amount of methods that report on this dataset is limited since it is recently released.

\subsection{Implementation Details and Evaluation}
% {\bf Implementation Details}
During training, the input images are re-sized to $384 \times 128$ and then augmented by random horizontal flip, normalization, and random erasing \cite{zhong2017random}. The testing images are re-sized to $384 \times 128$ and augmented only by normalization. In our experiments, the sizes of feature maps $\textbf{T}_a$ and $\textbf{T}_g$ are $1024\times24\times8$, and $2048\times24\times8$, respectively. We set the dimension of features ($k_a$, $k_g$) after global average-pooling both equal to 1024, leading to a 2048-dimensional final feature embedding for matching.

With the ImageNet-pretrained ResNet-50 backbone, we used the two-step transfer learning algorithm \cite{geng2016deep} to fine-tune the model. First, we freeze the backbone weights and only train the reduction layers, classifiers and all attention modules for 10 epochs with only the cross entropy loss and triplet loss applied. Second, all layers are freed for training for another 60 epochs, with the full loss (\ref{loss}) applied. We set $\beta_{tr}=10^{-1}$, $\beta_{OF}=10^{-6}$ and $\beta_{OW}=10^{-3}$, and the margin parameter for triplet loss $\alpha=1.2$. 
%. In this stage, cross entropy loss, triplet loss, and OW penalty were applied. Afterward, all four terms in loss are applied, and the model was trained for another 60 epochs. 

%We increased the parameter $p$ in dropout layers as:

%$$p = min(0.5, 0.2 + \lfloor epoch / 10 \rfloor * 0.1)$$

%while during training stage 1, $p$ is fixed to 0.2. We set $\beta_{tr}=10^{-1}$, $\beta_{OF}=10^{-6}$ and $\beta_{OW}=10^{-3}$, and the margin parameter for triplet loss $\alpha=1.2$. 

Our network is trained using 2 Tesla P100 GPUs with a batch size of 64. Each batch contains 16 identities, with 4 instances per identity. We use the Adam optimizer with the base learning rate initialized to $3\times 10^{-4}$, then decayed to $3\times 10^{-5}$,  $3\times 10^{-6}$ after 30, 40 epochs, respectively. The training takes about 4 hours on the Market-1501 dataset.

%{\bf Evaluation metrics} 
We adopt standard Re-ID metrics: top-1 accuracy, and the mean Average Precision (mAP). We consider mAP to be a more reliable indicator for Re-ID performance.

\subsection{Ablation Study of ABD-Net}
To verify the effects of attention modules and orthogonality regularization in ABD-Net, we incrementally evaluate each module on Market-1501 and DukeMTMC-Re-ID. We choose ResNet-50 \footnote{For the fairness of ablation study, we use two duplicated branches with the same res\_conv\_5 like the structure in ABD-Net as shown in Fig.～\ref{fig:arch}. Data augmentation and dropout are applied.} with the cross entropy loss (XE) as the baseline. 
% We built nine variants for the ablation study: a) basel. (XE) + O.F. , b) basel. (XE) + O.W. , c) basel. (XE) + O.F. + O.W. , d) basel. (XE) + PAM, e) basel. (XE) + CAM, f) basel. (XE) + PAM + CAM, g) ABD-Net (XE), h) ABD-Net (triplet) and i) ABD-Net (FAT). 
Nine variants are then constructed on top of the baseline\footnote{Note that (1) CAM is used in two places of ABD-Net; (2) ABD-Net adopts O.F. + O.W. + PAM + CAM.}: \textbf{a)} baseline (XE) + PAM; \textbf{b)} baseline (XE) + CAM; \textbf{c)} baseline (XE) + PAM + CAM; \textbf{d)} baseline (XE) + O.F.; \textbf{e)} baseline (XE) + O.W.; \textbf{f)} baseline  (XE) + O.F. + O.W.; \textbf{g)} baseline + SVD layer (similar to SVD-Net \cite{Sun_2017}); \textbf{h)} ABD-Net (XE), that sets $\beta_{tr}= 0$ in (\ref{loss}); and \textbf{i)} ABD-Net, that uses the full loss (\ref{loss}).

%The model denoted as (FAT) adopt fat loss assisted with cross entropy loss.

\begin{table}[ht]
\begin{center}
\caption{Ablation Study of ABD-Net on Market-1501. O.F. and O.W.: Orthogonality Regularization on Features and Weights; PAM and CAM: Position and Channel Attention Modules.}
\vspace{0.5em}
\label{table:ablation}
\resizebox{0.47\textwidth}{!}{
\begin{tabular}{l|c|c|c|c}
\hline
\multirow{2}{*}{Method} & \multicolumn{2}{c|}{Market-1501} & \multicolumn{2}{c}{DukeMTMC}\\ \cline{2-5} 
& top1 & mAP & top1 & mAP\\ \hline\hline
baseline (XE) & 91.50 & 77.40 & 82.80 & 66.40\\
\hline
baseline (XE) + PAM & 92.10 & 78.10 & 83.80 & 67.00 \\
baseline (XE) + CAM & 91.80 & 78.00 & 84.30 & 67.60 \\
baseline (XE) + PAM + CAM & 92.70 & 78.50 & 84.40 & 67.90\\
\hline
baseline (XE) + O.F. & 92.90 & 82.10 & 84.90 & 71.30\\
baseline (XE) + O.W. & 92.50 & 78.50 & 83.70 & 67.40\\
baseline (XE) + O.F. + O.W. & 93.20 & 82.30 & 85.30 & 72.20\\
baseline + SVD layer & 90.80 & 76.90 & 79.40 & 62.50\\
\hline
%basel. ($l^{(0)}$) + ABD module (a) & 92.40 & 80.00 & 85.20 & 68.60\\
%basel. ($l^{(0)}$) + ABD module (b) & 93.80 & 83.10 & 85.80 & 72.30\\
%\hline
ABD-Net (XE) & 94.90 & 85.90 & 87.30 & 76.00\\
\hline
ABD-Net & 95.60 & 88.28 & 89.00 & 78.59\\
% \hline
% ABD-Net (FAT) & & & & \\
\hline
\end{tabular}}
\end{center}
\vspace{-5mm}
\end{table}

Table~\ref{table:ablation} presents the ablation study results, from which several observations could be drawn:
\begin{itemize}
    \item Using either PAM or CAM improves the baseline on both datasets. The combination of the two different attention mechanisms gains further improvements, demonstrating their complementary power over utilizing either alone.
    
    \item Using either O.F. or O.W. consistently outperforms the baseline on both datasets, and their combination leads to further gains which validates the effectiveness of our orthogonality regularizations. We also observe that the proposed SVDO-based O.W. empirically performs better than the SVD layer, 
    %\footnote{Our results of baseline + SVD layer is better than SVDNet \cite{Sun_2017,zhong2017random}, but worse than baseline (XE). A possible reason is that SVD layer is a ``hard constraint" so that it restricts the backbone learning capability.}, 
    potentially because SVD layer acts as a ``hard constraint" and hence restricts the learning capability of the ResNet-50 backbone.
    \item By combining ``attention'' and ``diversity'', ABD-Net (XE) sees further boosts. For example, on Market-1501, ABD-Net (XE) outperforms the ``no attention'' counterpart (baseline (XE) + O.F. + O.W.) by a margin of 1.50\% (top-1)/3.60\% (mAP), and it outperforms ``no diversity'' counterpart (baseline (XE) + O.F. + O.W.) by 2.20\% (top-1)/7.40\% (mAP). Moreover, there are further performance improvements when we enforce diversity in the attention mechanism. 
    %As shown in Table~\ref{table:ablation}, the performance improvement of ABD-Net is larger than simply stacking these two components. 
    Finally, the full ABD-Net further benefits from adding triplet loss.
\end{itemize}

\subsection{Comparison to State-of-the-art Methods}
We compare ABD-Net against the state-of-the-art methods on Market-1501, DukeMTMC-Re-ID and MSMT17, as shown in Tables \ref{table:sota_market}, \ref{table:sota_duke}, and \ref{table:sota_msmt}, respectively. For fair comparison, no post-processing such as re-ranking \cite{zhong2017re} or multi-query fusion \cite{zheng2015scalable} was used for our methods. 

\begin{table}[ht]
\begin{center}
\caption{Comparison to state-of-the-art methods on Market-1501. {\color{red}Red} denotes our performance, and {\color{blue}Blue} denotes the best performance reported by existing methods: the same hereinafter.}
\vspace{0.5em}
\label{table:sota_market}
\begin{threeparttable}
\resizebox{0.44\textwidth}{!}{
\begin{tabular}{c|c|c}
\hline
\multirow{2}{*}{Method} & \multicolumn{2}{c}{Market-1501} \\ \cline{2-3} 
 & top1 & mAP \\ \hline\hline
BOW \cite{zheng2015scalable} (2015 ICCV) & 44.42 & 20.76  \\
Re-Rank \cite{zhong2017re} (2017 CVPR) & 77.11 & 63.63 \\
SSM \cite{bai2017scalable} (2017 CVPR) & 82.21 & 68.80 \\
\hline
\hline
SVDNet(RE) \cite{zhong2017random} (2017 CVPR) & 87.08 & 71.31 \\
AWTL \cite{ristani2018features} (2018 CVPR) & 84.20 & 68.03 \\
DSR \cite{he2018deep} (2018 CVPR) & 83.68 & 64.25 \\
MLFN \cite{chang2018multi} (2018 CVPR) & 90.00 & 74.30 \\
Deep CRF \cite{chen2018group} (2018 CVPR) & 93.50 & 81.60 \\
Deep KPM \cite{shen2018end} (2018 CVPR) & 90.10 & 75.30 \\
HAP2S \cite{yu2018hard} (2018 ECCV) & 84.20 & 69.76 \\
SGGNN \cite{shen2018person} (2018 ECCV) & 92.30 & 82.08 \\
Part-aligned \cite{suh2018part} (2018 ECCV) & 91.70 & 79.60 \\
PCB \cite{sun2018beyond} (2018 ECCV) & 93.80 & 81.60 \\
SNL \cite{li2018support} (2018 ACM MM) & 88.27 & 73.43 \\
HDLF \cite{zeng2018person} (2018 ACM MM) & 93.30 & 79.10 \\
$\ddagger$ MGN \cite{wang2018learning} (2018 ACM MM) & 95.70 & 86.90 \\
$\ddagger$ Local CNN \cite{yang2018local} (2018 ACM MM) & {\color{blue}95.90} & {\color{blue}87.40} \\
\hline
\hline
* MGCAM \cite{song2018mask} (2018 CVPR) & 83.79 & 74.33 \\ 
* AACN \cite{xu2018attention} (2018 CVPR) & 85.90 & 66.87 \\
* HA-CNN \cite{Li_2018} (2018 CVPR) & 91.20 & 75.70 \\
* C$A^3$Net \cite{Liu_2018} (2018 CVPR) & 93.20 & 80.00 \\
* Mancs \cite{wang2018mancs} (2018 ECCV) & 93.10 & 82.30\\
* $A^3$M \cite{han2018attribute} (2018 ACM MM) & 86.54 & 68.97 \\
\hline
\hline
$\bullet$ SPReID \cite{kalayeh2018human} (2018 CVPR) & 93.68 & 83.36 \\
$\ast$ $\diamond$ DuATM \cite{Si_2018} (2018 CVPR) & 91.42 & 76.62\\
\hline
\hline
\textbf{ABD-Net} & {\color{red}95.60} & {\color{red}88.28} \\
\hline
\end{tabular}}
\begin{tablenotes}
\footnotesize
\item[$\ast$] This also exploits attention mechanisms.
\item[$\bullet$] This is with a \textbf{ResNet-152} backbone.
\item[$\diamond$] This is with a \textbf{DenseNet-121} backbone.
\item[$\ddagger$] Official codes are not released. We report the numbers in the original paper, which are better than our re-implementation. 
%For MGN, the best re-implement results online is top1: \textbf{94.95}, mAP: \textbf{86.15} from \url{https://github.com/GNAYUOHZ/ReID-MGN}.
\end{tablenotes}
\end{threeparttable}
\end{center}
\vspace{-4mm}
\end{table}

\begin{table}[ht]
\begin{center}
\caption{Comparison to state-of-the-art methods on DukeMTMC.}
\vspace{-2.5mm}
\label{table:sota_duke}
\begin{threeparttable}
\resizebox{0.44\textwidth}{!}{
\begin{tabular}{c|c|c}
\hline
\multirow{2}{*}{Method} & \multicolumn{2}{c}{DukeMTMC-Re-ID} \\ \cline{2-3} 
 & top1 & mAP \\ \hline\hline
BOW \cite{zheng2015scalable} (2015 ICCV) & 25.13 & 12.17 \\
\hline
\hline
SVDNet \cite{Sun_2017} (2017 ICCV) & 76.70 & 56.80 \\
SVDNet(RE) \cite{zhong2017random} (2017 CVPR) & 79.31 & 62.44 \\
FMN \cite{ding2017let} (2017 CVPR) & 74.51 & 56.88 \\
PAN \cite{zheng2017pedestrian} (2018 TCSVT) & 71.59 & 51.51 \\
AWTL(2-stream) \cite{ristani2018features} (2018 CVPR) & 79.80 & 63.40 \\
Deep-person \cite{bai2017deepperson} (2018 CVPR) & 80.90 & 64.80 \\
MLFN \cite{chang2018multi} (2018 CVPR) & 81.20 & 62.80 \\
GP-Re-ID \cite{almazan2018reid} (2018 CVPR) & 85.20 & 72.80\\
PCB \cite{sun2018beyond} (2018 ECCV) & 83.30 & 69.20 \\
Part-aligned \cite{suh2018part} (2018 ECCV) & 84.40 & 69.30 \\
$\ddagger$ MGN \cite{wang2018learning} (2018 ACM MM) & {\color{blue}88.70} & {\color{blue}78.40} \\
$\ddagger$ Local CNN \cite{yang2018local} (2018 ACM MM) & 82.23 & 66.04 \\
\hline
\hline
* AACN \cite{xu2018attention} (2018 CVPR) & 76.84 & 59.25 \\
* HA-CNN \cite{Li_2018} (2018 CVPR) & 80.50 & 63.80 \\
* C$A^3$Net \cite{Liu_2018} (2018 CVPR) & 84.60 & 70.20 \\
* Mancs \cite{wang2018mancs} (2018 ECCV) & 84.90 & 71.80\\
\hline
\hline
$\bullet$ SPReID \cite{kalayeh2018human} (2018 CVPR) & {85.95} & {73.34} \\
$\ast$ $\diamond$ DuATM \cite{Si_2018} (2018 CVPR) & 78.74 & 62.26\\
\hline
\hline
\textbf{ABD-Net} & {\color{red}89.00} & {\color{red}78.59} \\
\hline
\end{tabular}}
\begin{tablenotes}
\footnotesize
\item[$\ast$] This also exploits attention mechanisms.
\item[$\bullet$] This is with a \textbf{ResNet-152} backbone.
\item[$\diamond$] This is with a \textbf{DenseNet-121} backbone.
\item[$\ddagger$] Official codes are not released. We report the numbers in the \\ original paper, which are better than our re-implementation. 
\end{tablenotes}
\end{threeparttable}
\end{center}
\vspace{-3mm}
\end{table}

\begin{table}[ht]
\begin{center}
\caption{Comparison to state-of-the-art methods on MSMT17.}
\vspace{0.5em}
\label{table:sota_msmt}
\resizebox{0.4\textwidth}{!}{
\begin{tabular}{c|c|c|c}
\hline
\multirow{2}{*}{Method} & \multicolumn{3}{c}{MSMT17} \\ \cline{2-4} 
 & top1 & top5 & mAP \\ \hline\hline
PDC \cite{su2017pose} (2017 ICCV) & 58.00 & 73.60 & 29.70 \\
GLAD \cite{wei2017glad} (2017 ACM MM) & {\color{blue}61.40} & {\color{blue}76.80} & {\color{blue}34.00} \\
\hline
\hline
\textbf{ABD-Net} & {\color{red}82.30} & {\color{red}90.60} & {\color{red}60.80} \\
\hline
\end{tabular}}
\end{center}
\vspace{-2mm}
\end{table}

ABD-Net has clearly yielded overall state-of-the-art performance on all datasets. Specifically, on DukeMTMC-Re-ID, ABD-Net obtains $89.00\%$ top-1 accuracy and $78.59\%$ mAP, which significantly outperforms all existing methods. On MSMT17, ABD-Net presents a clear winner case too. On Market-1501, its top-1 accuracy (95.60\%) slightly lags behind Local CNN \cite{yang2018local}  (95.90\%) and MGN \cite{wang2018learning} (95.70\%); yet ABD-Net clearly surpasses all existing methods in terms of mAP (88.28\%, outperforming the closest competitor \cite{yang2018local} by a large margin of 0.88\%).

Specifically, we emphasize the comparison between ABD-Net and existing attention-based methods (marked by $\ast$ in the Tables \ref{table:sota_market}  \ref{table:sota_duke}). As shown in Table \ref{table:sota_market} and \ref{table:sota_duke}, ABD-Net achieves at least $2.40\%$ top-1 and $5.98$\% mAP improvement on Market-1501, compared to the closest attention-based prior work C$A^3$Net \cite{Liu_2018}. On DukeMTMC, the margin becomes $3.40\%$ for top-1 and $6.40$\% for mAP. We also considered SVDNet \cite{Sun_2017} and HA-CNN \cite{Li_2018} which also proposed to generate diverse and uncorrelated feature embeddings. ABD-Net surpasses both with significant top-1 and mAP improvement. Overall, our observations endorse the superiority of ABD-Net by combing ``attentive'' and ``diverse''. 

%as well as $3.40\%$ top-1 improvement and $6.40\%$ mAP improvement on DukeMTMC-Re-ID. The results suggest that the highly correlated attentive feature embedding from most of previous attention mechanisms are not enough representative. In our proposed ABD-Net, enforcing the attentive feature embedding to be more diverse through the orthogonality regularization leads to a more representative, robust and discriminative feature space, which further enhances the person Re-ID performance.

%Among these previous work in Table \ref{table:sota_market} and Table \ref{table:sota_duke}, there are two methods (SVDNet and HA-CNN) which also focusing on generating more diverse and representative feature embedding. SVDNet performs SVD on the weight matrix of FC layer to reduce the redundancy in the FC descriptor and enforce the diverse of weight vectors. Instead, ABD-Net not only apply the orthogonality regularization on weight but also directly constrain the diversity of feature embedding. It's no wonder that ABD-Net surpass SVDNet as least 8.6\% of top-1 accuracy and 15.7\% of mAP on Market-1501 and DukeMTMC-Re-ID. HA-CNN selects attention regions from multi-granularity feature maps and add them together to learn an attentive and \textbf{handcrafted} diverse feature embedding. Here, from the Table \ref{table:sota_market} and Table \ref{table:sota_duke}, our proposed learned attentive but diverse feature embedding from ABD-Net outperforms HA-CNN with significant top-1 and mAP advantages. 

\subsection{Visualizations}
% \subsection{Visualizations\footnote{To fairly evaluate the contribution of our proposed attentive mechanism and diversity regularization, we exclude the effect of triplet loss, and only compare the following three methods: the baseline (XE), baseline (XE) + PAM + CAM, and ABD-Net (XE).}} \label{sec:vis}
% \textbf{Visualization of Attentive but Diverse Module.}
% \footnote{Here we did not use our best model ABD-Net, since we hope to ensure the fair comparison of the three methods, using the same XE loss.}

\begin{figure}[t]
\begin{center}
   \includegraphics[width=1\linewidth]{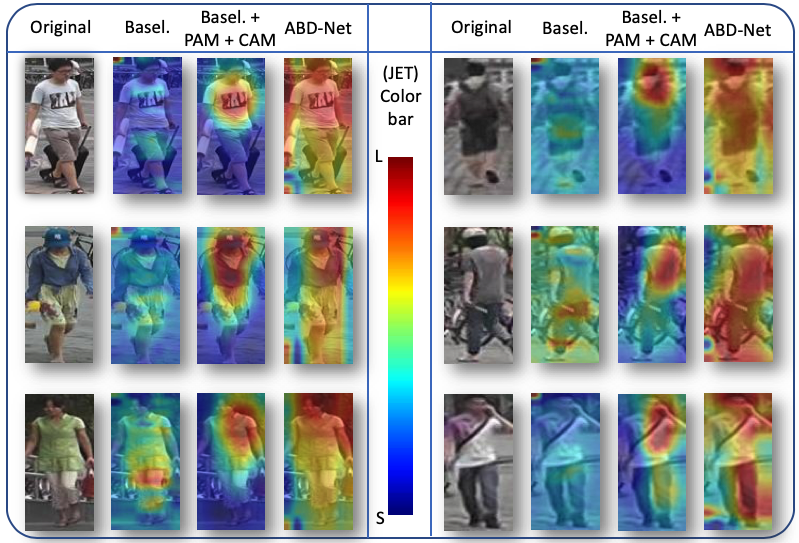}
\end{center}
   \caption{Visualization of attention maps from Baseline, Baseline + PAM + CAM and ABD-Net (XE). As shown in column four and eight, the diverse attention map from ABD-Net almost span over the whole person rather than overfit in some local regions.}
\label{fig:visf}
\vspace{-1mm}
\end{figure}

\paragraph{Attention Pattern Visualization:} We conduct a set of attention visualizations\footnote{Grad-CAM visualization method \cite{Selvaraju_2017}: \url{https://github.com/utkuozbulak/pytorch-cnn-visualizations}; RAM visualization method \cite{Yang_2019_CVPR} for testing images. More results can be found in the supplementary.} on the final output feature maps of the baseline (XE), baseline (XE) + PAM + CAM, and ABD-Net (XE), as shown in Fig.～\ref{fig:visf}. We notice that the feature maps from the baseline show little attentiveness. PAM + CAM enforces the network to focus more on the person region, but the attention regions can sometimes overly emphasize some local regions (\textit{e.g.}, clothes), implying the risk of overfitting person-irrelevant nuisances. Most channels focus on the similar region may also cause a high correlation in the feature embeddings. In contrast, the attention of ABD-Net (XE) can strike a better balance: it focuses on more of the local parts of the person's body while still being able to eliminate the person from backgrounds. The attention patterns now differ more from person to person, and the feature embeddings become more decorrelated and diverse. 

%focus on different parts of the person, indicating that it decorrelates the feature embedding and improve the diversity of feature embedding.

\begin{figure}[t]
\begin{center}
   \includegraphics[width=0.9\linewidth]{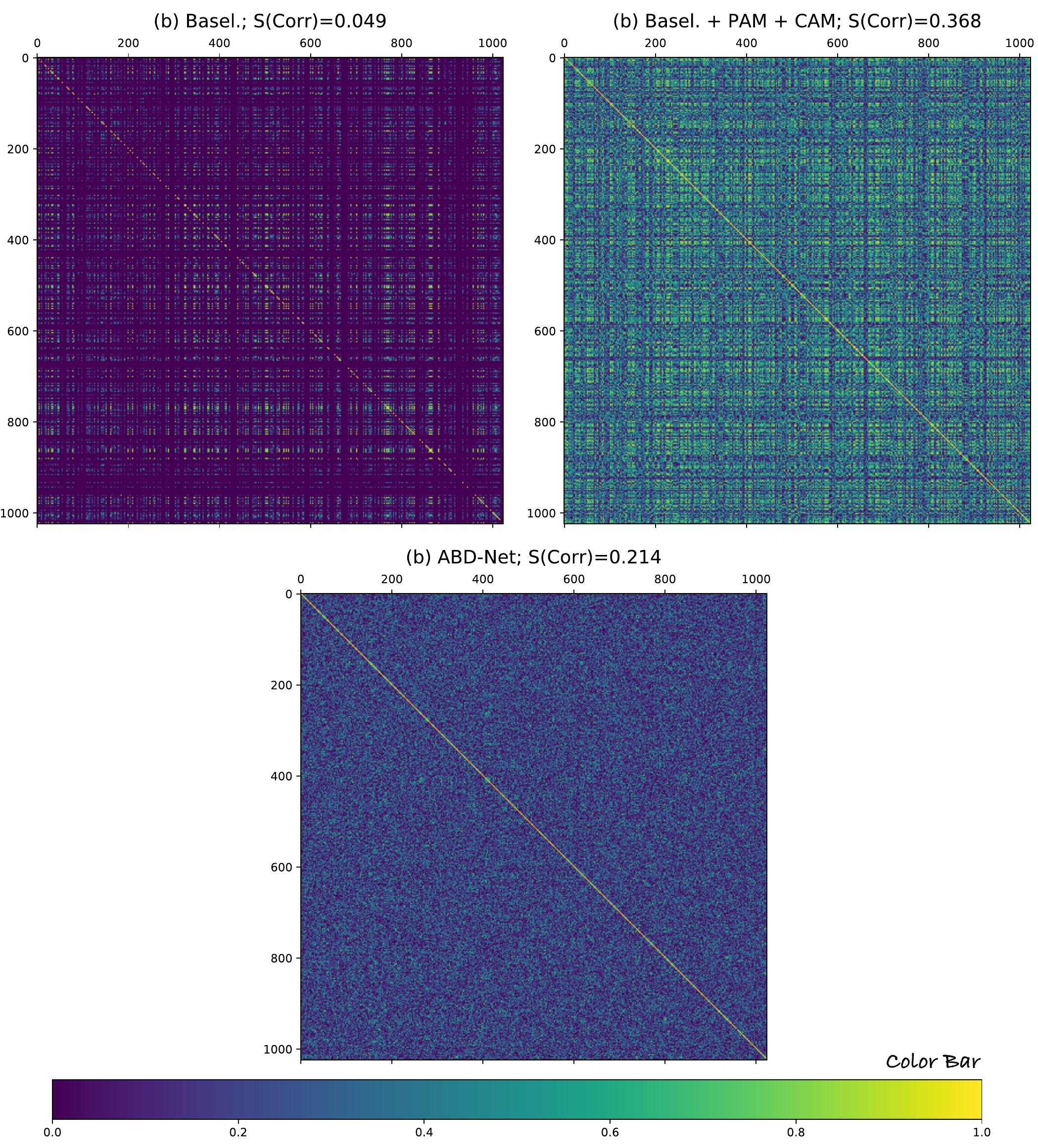}
\end{center}
\vspace{-1mm}
   \caption{Visualization of correlation matrix between channels from Baseline, Baseline + PAM + CAM and ABD-Net (XE). Brighter color indicates larger correlation. In clockwise order from top left image, attention brings feature embeddings high correlation, diversity reduces the redundancy and further improve the discriminative.}
\label{fig:visc}
\vspace{-1mm}
\end{figure}

\begin{figure}[t]
\begin{center}
   \includegraphics[width=1\linewidth]{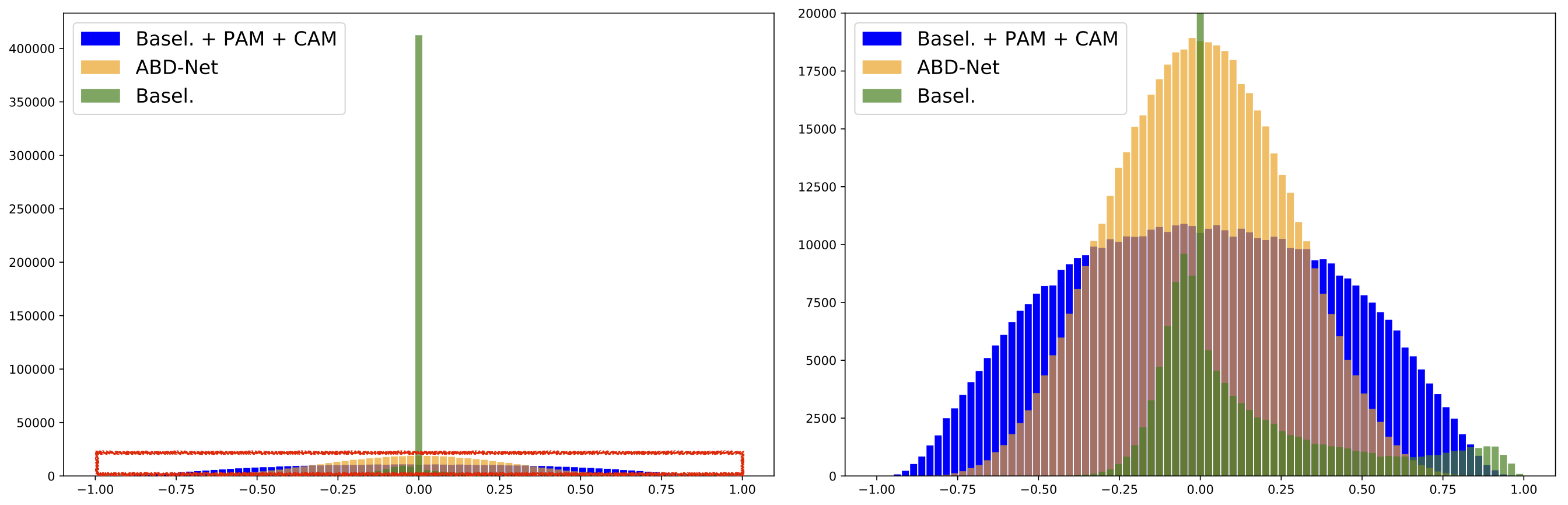}
\end{center}
   \caption{Histogram of correlation from Baseline, Baseline + PAM + CAM and ABD-Net (XE). A more skewed distribution indicates better de-correlated feature embeddings. (b) is a zoom-in view of the red box area in (a).}
\label{fig:visch}
\vspace{-1mm}
\end{figure}

\paragraph{Feature De-correlation:} We study the correlation matrix between the channel outputs produced by Baseline, Baseline + PAM + CAM and ABD-Net (XE) \footnote{Here we used a random testing image as the example and we offer more results in the supplementary.}. The feature embedding before the global average pooling is 
% $\textbf{F}\in \mathbb{R}^{C\times H\times W}$ and we 
reshaped into $\textbf{F}\in \mathbb{R}^{C\times N}$, where $N=H\times W$. Then, we visualize the correlation coefficient matrix for $\textbf{F}$, denoted as $\textbf{Corr}\in\mathbb{R}^{C\times C}$\footnote{We take the absolute value for correlation coefficients.} in Fig.～\ref{fig:visc}, and also compute the average of all correlation coefficients in each setting. The baseline feature embeddings reveal low correlations (\textbf{0.049} in average) in off-diagonal elements. After applying PAM and CAM, the feature correlations become much larger (\textbf{0.368} in average), supporting our hypothesis that the attention mechanism tends to encourage more ``focused'' and thus highly correlated features. However, with our orthogonality regularization, the feature correlations in ABD-Net (XE) are successfully suppressed (\textbf{0.214} in average) compared to the attention-only case. The feature histogram plots in Fig.～\ref{fig:visch} also certify the same observation.

\begin{figure}[t]
\begin{center}
   \includegraphics[width=1\linewidth]{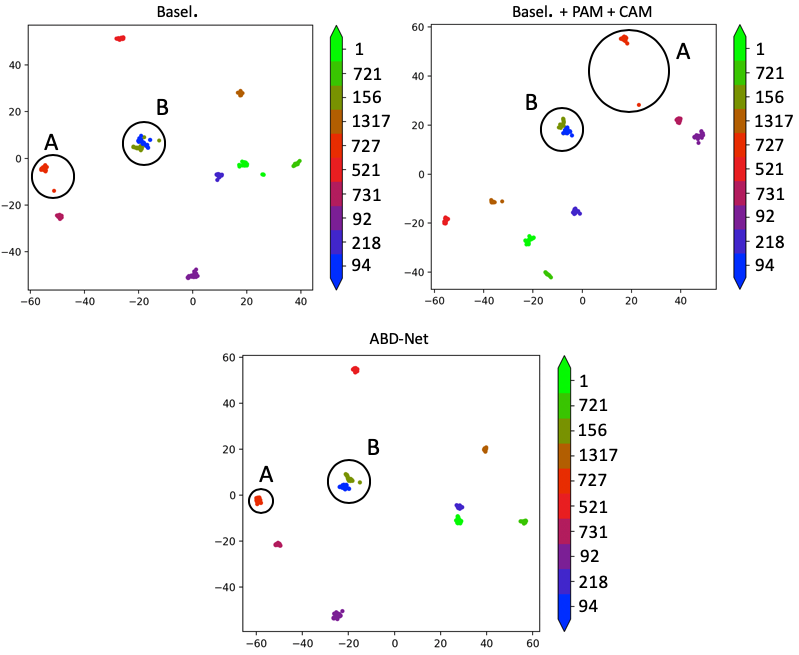}
\end{center}
   \caption{t-SNE visualization of feature distributions, from Baseline, Baseline + PAM + CAM and ABD-Net (XE). Ten identities are randomly selected from the Market-1501 and their IDs are listed on the right side of graphs. Circles A and B contain the features from IDs 521, 94 and 156, respectively.}
\label{fig:tsne}
\vspace{-1mm}
\end{figure}

\paragraph{Feature Embeddings Distributions:} Fig.～\ref{fig:tsne} shows the t-SNE visualization on feature distributions from Baseline, Baseline + PAM + CAM and ABD-Net (XE) using t-SNE. Compared with Baseline, although attentive features from Baseline + PAM + CAM make ID 94 and ID 156 in cycle B slightly distinguishable, ABD-Net enlarges the intra-class distance of ID 521 in cycle A. It makes the features from ID 94 and ID 156 more discriminative, meanwhile the features from ID 521 also lie in a compact region.

\begin{figure}[t]
\begin{center}
   \includegraphics[width=0.95\linewidth]{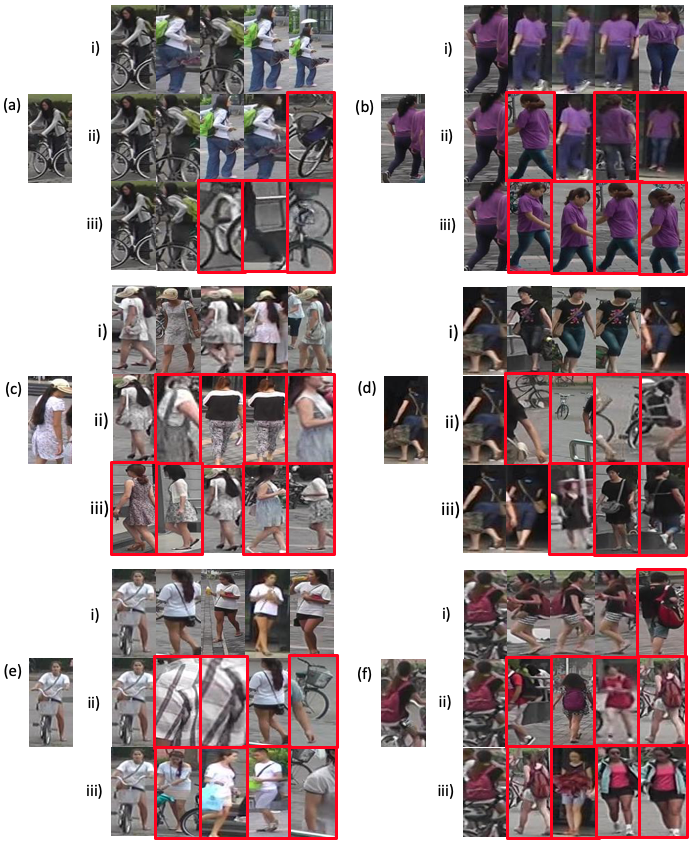}
\end{center}
   \caption{Six Re-ID examples of ABD-Net (XE), Baseline + PAM + CAM and Baseline on Market-1501. Left: query image. Right: i): top-5 results of ABD-Net (XE). ii): top-5 results of Baseline + PAM + CAM. iii): top-5 results of Baseline. Images in red boxes are negative results. Attentive but diverse feature embeddings boost the retrieval preformance.}
\label{fig:top5}
\vspace{-1mm}
\end{figure}

\paragraph{Re-ID Qualitative Visual Results:} Fig.～\ref{fig:top5} shows Re-ID visual examples of ABD-Net (XE), Baseline + PAM + CAM and Baseline on Market-1501. They indicate that ABD-Net succeeds in finding more true positives than Baseline + PAM + CAM model, even when the persons in the images are under significant view changes and appearance variations. %Fig.～\ref{fig:top5} (d) containts one failure case of ABD-Net, where the person with similar legs, cloth, hair, and bag are presented (it fails most competitior methods too.

\section{Conclusion} \label{sec:conclusion}
%\vspace{-3mm}
This paper proposes a novel Attentive but Diverse Network (ABD-Net) to learn more representative, robust, discriminative feature embeddings for person Re-ID.
%representative feature space for Person Re-ID. Our Attentive but Diverse paradigm can be applied to any existing architectures. The core components of proposed method ABD-Net are the attention module and the orthogonality regularization which simultaneously encourage the feature embedding to be attentive and diverse.
% ABD-Net is trained via a fat loss assisted with a cross-entropy loss.
ABD-Net demonstrates its state-of-the-art performance through extensive experiments where the ablations and visualizations show that each added component substantially contributes to its final performance. In the future, we will generalize the design concept of ABD-Net to other computer vision tasks. 

%conducted on three public person Re-ID datasets confirmed that our proposed framework achieves \textbf{the state-of-the-art} performance, and carefully-designed visualizations verified that the proposed attentive and diverse terms performed as expected.

{\small
\bibliographystyle{unsrt}
\bibliographystyle{ieee_fullname}
\bibliography{egbib}
}

\end{document}